\documentclass[conference]{IEEEtran}
\usepackage{times}

\usepackage[numbers]{natbib}
\usepackage{capt-of}
\usepackage{multicol}
\usepackage[bookmarks=true]{hyperref}
\usepackage[table]{xcolor}
\usepackage{algorithm}
\usepackage{algorithmic}
\usepackage{bm}
\usepackage{amsmath, amsfonts}
\usepackage{booktabs}
\usepackage{xcolor}
\usepackage{graphicx}
\usepackage{multirow}
\usepackage{caption} 
\usepackage{relsize}
\begin{document}

\title{X-Loco: Towards Generalist Humanoid Locomotion Control via Synergetic Policy Distillation}


\author{
    \authorblockN{
        Dewei Wang\textsuperscript{1, 2} \quad 
        Xinmiao Wang\textsuperscript{2, 3} \quad 
        Chenyun Zhang\textsuperscript{2} \quad 
        Jiyuan Shi\textsuperscript{2} \quad 
        Yingnan Zhao\textsuperscript{3} \quad \\
        Chenjia Bai\textsuperscript{2 *} \thanks{*Corresponding Author}
        Xuelong Li\textsuperscript{2 *}
    }
    \authorblockA{
        \textsuperscript{1}University of Science and Technology of China\quad
        \textsuperscript{2}Institute of Artificial Intelligence (TeleAI), China Telecom \\
        \textsuperscript{3}Harbin Engineering University\quad \\
    \textsuperscript{$*$}Corresponding author\quad
    Website: \href{https://x-loco-humanoid.github.io/}{\texttt{x-loco-humanoid.github.io}}
    }
}

\twocolumn[{%
\renewcommand\twocolumn[1][]{#1}%
\maketitle
\begin{center}
    \centering
    \captionsetup{type=figure}
     \includegraphics[width=1.0\textwidth]{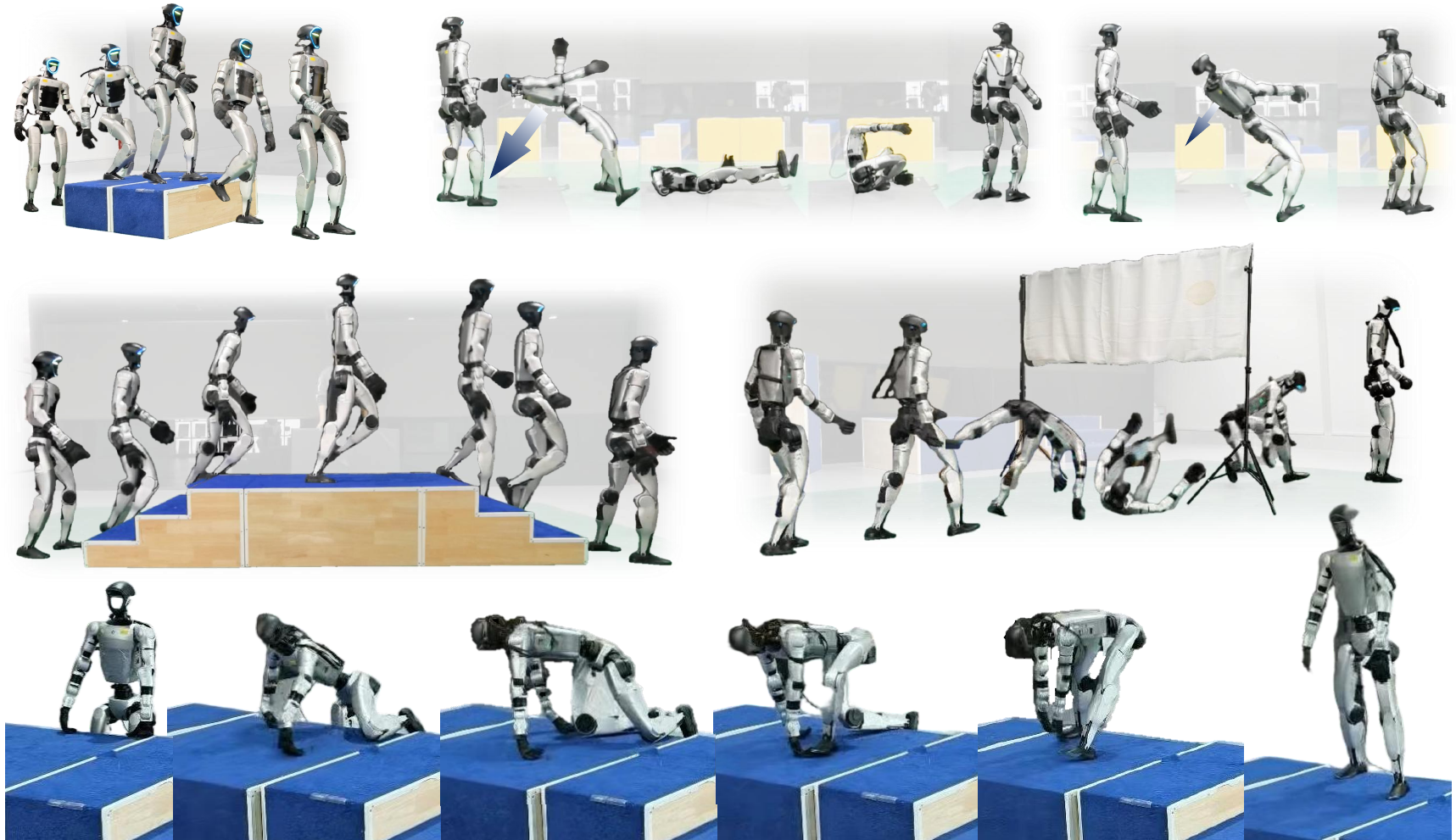}
     \vspace{-0.0in}
    \caption{X-Loco achieves vision-based generalist humanoid locomotion control. Relying solely on velocity commands without reference motions, X-Loco leverages proprioception and exteroception to traverse complex terrains, performing behaviors such as climbing up and down stairs, fall recovery, box climbing, and forward rolls.} 
    \label{fig:cover}
\end{center}
}]

\begin{abstract}
While recent advances have demonstrated strong performance in individual humanoid skills such as upright locomotion, fall recovery and whole-body coordination, learning a single policy that masters all these skills remains challenging due to the diverse dynamics and conflicting control objectives involved. To address this, we introduce X-Loco, a framework for training a vision-based generalist humanoid locomotion policy. X-Loco trains multiple oracle specialist policies and adopts a synergetic policy distillation with a case-adaptive specialist selection mechanism, which dynamically leverages multiple specialist policies to guide a vision-based student policy. This design enables the student to acquire a broad spectrum of locomotion skills, ranging from fall recovery to terrain traversal and whole-body coordination skills. To the best of our knowledge, X-Loco is the first framework to demonstrate vision-based humanoid locomotion that jointly integrates upright locomotion, whole-body coordination and fall recovery, while operating solely under velocity commands without relying on reference motions.
Experimental results show that X-Loco achieves superior performance, demonstrated by tasks such as fall recovery and terrain traversal. Ablation studies further highlight that our framework effectively leverages specialist expertise and enhances learning efficiency.
\end{abstract}

\IEEEpeerreviewmaketitle

\section{Introduction}
Humanoid robots equipped with advanced algorithms have garnered significant attention due to their potential for anthropomorphic motion and versatile manipulation~\cite{tong2024advancements, darvish2023teleoperation, qu2025spatialvla, billard2019trends, song2025hume, asfour2008toward, hm:he2025asap, an2026ai}. As a fundamental capability, locomotion control has witnessed rapid progress facilitated by reinforcement learning (RL) and high-fidelity physical simulators~\cite{sutton1998introduction, schulman2017proximal, hl:gu2024humanoid, mittal2025isaac, makoviychuk2021isaac, zakka2025mujoco, hr:huang2025learning, hm:he2025asap}.
Recent advancements have demonstrated remarkable results, ranging from traversing challenging terrains~\cite{hl:wang2025more, hl:long2025learning, hl:song2025gait, hl:gu2024advancing} and executing robust fall recovery~\cite{hr:chen2025hifar, hr:he2025learning, hr:xu2025unified} to agilely tracking highly dynamic motions~\cite{hm:he2025asap, hm:liao2025beyondmimic, hm:yin2025unitracker, hm:xie2025kungfubot, hm:han2025kungfubot2, hm:zhang2025track}.

Despite these advancements, a key limitation of many existing humanoid locomotion control methods lies in their predominant focus on isolated or specific categories of locomotion skills, such as exteroception-based terrain traversal~\cite{hl:wang2025more, hl:long2025learning, hl:song2025gait, hl:zhuang2024humanoid} or fall recovery~\cite{hr:huang2025learning, hr:xu2025unified, luo2014multi, stuckler2006getting}. This functional fragmentation renders these methods incapable of handling complex scenarios, such as autonomously resuming locomotion after a fall. Furthermore, these methods fail to achieve whole-body coordination skills, which prevents the robot from traversing challenging terrains. While motion tracking~\cite{hm:han2025kungfubot2, hm:yin2025visualmimic, hm:zeng2025behavior, hm:yin2025unitracker, hm:yang2025omniretarget} and teleoperation paradigms~\cite{ht:omnih2o, ht:ze2025twist, ht:ji2025exbody2advancedexpressivehumanoid} facilitate diverse whole-body coordination behaviors, their inference relies heavily on reference motions or human intervention via specialized devices, leaving the robot unable to autonomously perceive its surroundings and perform locomotion. Consequently, these approaches lack the necessary autonomy required for self-contained and versatile humanoid locomotion controllers.

Developing a vision-based generalist locomotion controller capable of integrating locomotion skills ranging from terrain traversal to fall recovery, alongside contact-rich skills like box climbing, remains an open challenge in humanoid robotics. To realize such a versatile controller, several challenges must be addressed: 
First, reward engineering for humanoid locomotion remains inherently labor-intensive~\cite{hr:huang2025learning, hl:xue2025unified}, as formulating reward functions capable of eliciting diverse motor skills poses a significant challenge.
Second, achieving whole-body coordination skills~\cite{hm:yang2025omniretarget, hm:han2025kungfubot2} without reference motion introduces substantial complexity, as the absence of guidance leads to inefficient exploration within high-dimensional state-action spaces.
Third, beyond mastering diverse skills, the controller needs to leverage visual perception to understand the environment and generate appropriate, real-time responses to conditions. 
Lastly, while head-mounted cameras facilitate perception, camera rendering in current simulations is limited by slow or non-independent rendering across parallel environments. Achieving fast and decoupled camera rendering is essential for the efficient training of vision-based policies.

To address these challenges, we introduce X-Loco, a framework that develops a generalist humanoid controller by distilling three privileged specialist policies, each optimized for a distinct capability: upright locomotion, fall recovery, and whole-body coordination, such as rolling and box climbing. Specifically, we implement a \emph{Case-Adaptive Specialist Selection} (CASS) mechanism that dynamically queries the most relevant specialist policy based on the robot's state and the surrounding terrains giving the action guidance for the student policy. As acquiring multiple locomotion skills simultaneously is often hindered by the untrained policy inability to cover the specialists' optimal state-action distribution, we introduce \emph{Specialist Annealing Rollout} (SAR), a dynamic ratio-based data collection strategy that incorporates a proportion of rollouts from the specialist policies for training. This mixing ratio decays as the distillation loss converges, shifting the focus toward the student policy's own explorations while reducing noisy data in the early stages of training. To enhance robustness, \emph{Stochastic Fall Injection} (SFI) is implemented by applying active external disturbances during locomotion, rather than just initializing fallen states. This mechanism forces the policy to adapt to unexpected loss of balance, requiring the emergence of a transition between locomotion and emergency fall recovery.

In summary, by distilling specialist policy expertise into a generalist policy, X-Loco expands the capabilities of existing humanoid locomotion controllers.
Our primary contributions are summarized as follows:
\begin{itemize}
    \item We develop X-Loco, which trains three locomotion specialist policies sharing a unified action space and integrates their diverse motor skills into a generalist policy.
    
    \item We introduce a synergetic distillation paradigm to consolidate expertise from multiple specialist policies into a single generalist policy, effectively mitigating interference between diverse locomotion skills.

    \item We introduce SAR to ensure efficient expert knowledge internalization through adaptive rollout mixing, and SFI to expose the policy to failure regimes, thereby enabling transitions between locomotion and fall recovery.

    \item We deploy X-Loco on the Unitree G1 robot, demonstrating its superior performance and stability across diverse terrains and challenging scenarios, thereby validating the framework's robustness and sim-to-real transferability.
\end{itemize}

\begin{figure*}[t]
    \centering
    \includegraphics[width=1.0\linewidth]{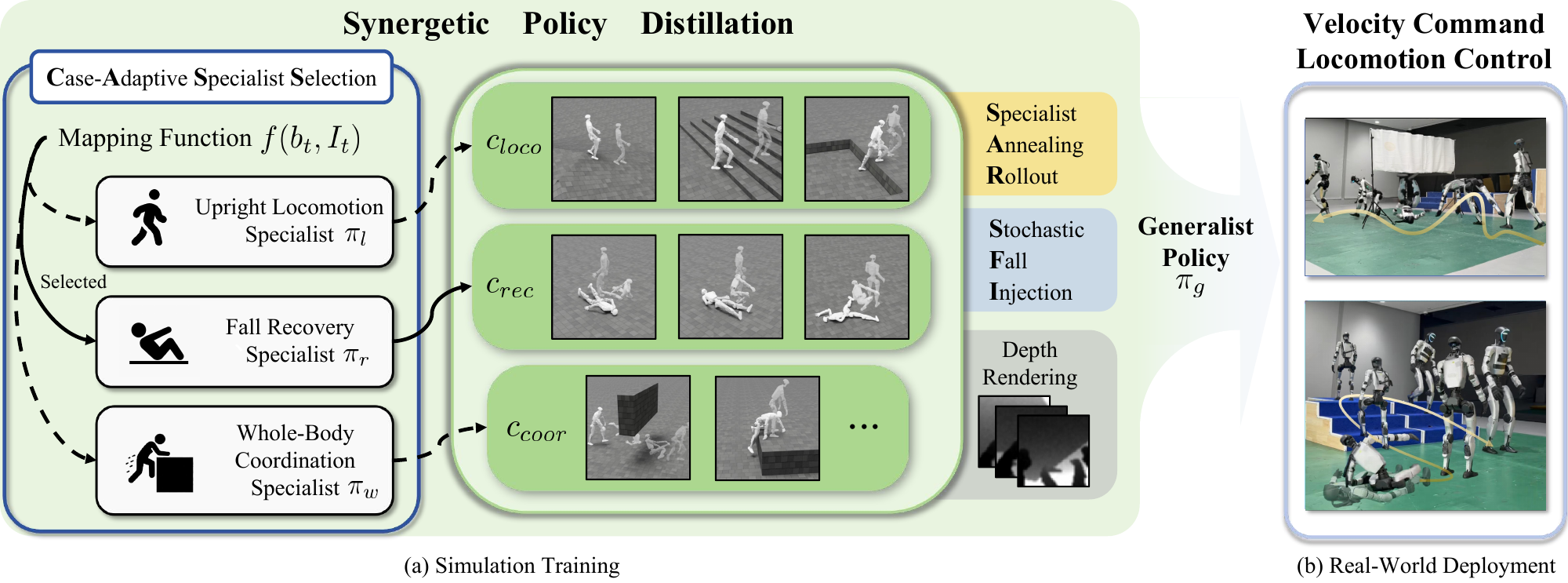}
    \vspace{-0.30cm}
    \caption{\textbf{Overview of X-Loco}. (a) X-Loco integrates the capabilities of three specialist policies into a vision-based generalist policy via Synergetic Policy Distillation. (b) X-Loco can perform diverse locomotion skills in the real world. }
    \label{fig:method}
\end{figure*}

\section{Related Work}

\subsection{Learning-based Humanoid Control}
\subsubsection{Upright Locomotion}
Humanoid locomotion has evolved from blind locomotion on uneven terrains~\cite{hl:gu2024humanoid, hl:radosavovic2024real, hl:zhang2024whole, hl:gu2024humanoid}, visual locomotion across complex terrains~\cite{hl:wang2025more, hl:song2025gait, hl:long2025learning} and multiple locomotion skill learning~\cite{hl:zhuang2024humanoid, hl:xue2025unified}. By decoupling the control of the upper and lower limbs~\cite{hl:ben2025homie, hl:shi2025adversarial, hl:xue2025unified}, humanoid robots can achieve complex loco-manipulation tasks that require coordination between the entire body. However, the aforementioned methods fall short of addressing complex tasks such as fall recovery and high-difficulty whole-body coordinated movements.

\subsubsection{Fall Recovery}
Falling is an inevitable occurrence for legged robots. RL provides a robust and generalizable framework for fall recovery~\cite{hr:jeong2016efficient, hr:huang2025learning, hr:chen2025hifar}. FIRM~\cite{hr:xu2025unified} has explored the synergy between fall-safety and fall-recovery and AHC~\cite{hl:zhao2025towards} utilizes multi-task RL to simultaneously achieve both autonomous fall recovery and stable walking. Despite these advancements, integrating fall recovery with a visual locomotion controller remains an open challenge.

\subsubsection{Motion Tracking}
Leveraging motion retargeting~\cite{hm:araujo2025retargeting, hm:Zakka_Mink_Python_inverse_2025, hm:he2024learning} and motion imitation~\cite{hm:peng2018deepmimic, hm:tessler2024maskedmimic, hm:luo2023perpetual}, humanoid robots can perform highly dynamic behaviors~\cite{hm:he2025asap, hm:xie2025kungfubot, hm:zhang2025track, hm:liao2025beyondmimic} and universal motion tracking~\cite{hm:han2025kungfubot2, hm:zeng2025behavior, hm:yin2025unitracker, hm:chen2025gmt} given reference motions. Exteroceptive information is integrated into a motion tracking framework to enable complex loco-manipulation~\cite{hm:yin2025visualmimic} and sensorimotor locomotion~\cite{hm:allshire2025visual}. Our method further eliminates the dependency on reference motions during inference, enabling the robot to autonomously perform whole-body coordination movements based on exteroceptive perception.

\subsection{Multiple Locomotion Skills Learning}
While developing a single policy for multiple skills is often hindered by gradient interference and task objective trade-offs~\cite{ms:kalashnikov2021mt, ms:yu2020gradient}. Mixture-of-Experts (MoE)~\cite{celik2024acquiring, jacobs1991adaptive} has been shown to mitigate gradient conflict in multi-skill learning and effectively facilitate the acquisition of diverse locomotion gaits~\cite{hl:wang2025more, ms:huang2025moe}. Methods such as MELA~\cite{ms:yang2020multi} and MTAC~\cite{ms:shah2023mtac} utilize a hierarchical structure to decouple individual motor skills, which are subsequently coordinated by a high-level module. Locomotion skills with minor morphological variations can be acquired by one-stage framework leveraging reward function design~\cite{hl:xue2025unified, margolis2023walk} and curriculum learning~\cite{cheng2024extreme, wang2025skill}.

In legged locomotion, policy distillation is widely utilized for privileged information learning~\cite{cheng2024extreme, ht:omnih2o, hm:chen2025gmt} and sim-to-real transfer~\cite{hm:yin2025visualmimic, kumar2021rma}. Policy distillation enables dynamic parkour behaviors by training specialized teachers for different terrains and distilling their expertise into a single policy~\cite{zhuang2023robot, hl:zhuang2024humanoid}. In contrast to existing works, our framework further integrates whole-body coordination and fall recovery into a generalist policy, autonomously executed based solely on exteroceptive perception and velocity commands.

\section{Problem Formulation} 
We formulate the humanoid locomotion control as a Markov Decision Process (MDP), defined by a tuple $\mathcal M=(\mathcal S, \mathcal A, \mathcal P, R, \gamma)$ where $\mathcal{S}$ and $\mathcal{A}$ denote the state and action spaces, respectively. $\mathcal{P}(\bm{s}_{t+1} | \bm{s}_t, \bm{a}_t)$ represents the state transition probability, $R(\bm{s}_t, \bm{a}_t): \mathcal{S} \times \mathcal{A} \rightarrow \mathbb{R}$ is the reward function, and $\gamma \in [0, 1)$ is the discount factor. We employ proximal policy optimization (PPO) \cite{schulman2017proximal} and Generalized Advantage Estimation (GAE) \cite{schulman2015high} to train specialist policies, each optimized to maximize the expected discounted cumulative return $J(\pi) = \mathbb{E}_{\tau \sim \pi} \left[ \sum_{t=0}^{\infty} \gamma^t R(\bm{s}_t, \bm{a}_t) \right]$. 

\subsubsection{State Space}
We partition the state space $\mathcal{S}$ into three components: the privileged state $\bm{s}^{\text{priv}}_t$, which contains information available only in simulation, proprioceptive information $\bm{s}^{\text{prop}}_t$ and exteroceptive information which in our setting consists of depth images $\bm{d}_t \in \mathbb{R}^{64 \times 64}$. The proprioceptive information $\bm{s}^{\text{prop}}_t$ primarily comprises:
\begin{equation}
\bm s_t^{\text{prop}} = [\bm{\omega}_t, \bm g_t, \bm q_t, \bm{\dot{q}}_t, \bm a_{t-1}] \in \mathbb{R}^{78},
\end{equation}
where $\bm{\omega}_t \in \mathbb{R}^3$ is the base angular velocity, $\bm g_t \in \mathbb{R}^3$ is the gravity vector in the base frame, $\bm q_t \in \mathbb{R}^{23}$ and $\bm{\dot{q}_t} \in \mathbb{R}^{23}$ represent joint position and joint velocity, and $\bm a_{t-1} \in \mathbb{R}^{23}$ is the last action. The privileged state $\bm{s}^{\text{priv}}_t$ is defined as: 
\begin{equation}
\bm s_t^{\text{priv}} = [\bm{v}_t, b_t, \bm{e}_t, \bm f_t, \bm h_t, \bm{m}_t],
\end{equation}
where $\bm{v}_t \in \mathbb{R}^3$ is the base linear velocity, $b_t$ denotes the head height, and $\bm{e}_t \in \mathbb{R}^6$, $\bm f_t \in \mathbb{R}^6$ represent the positions and velocities of the hands and feet relative to the base frame, respectively. $\bm h_t \in \mathbb{R}^{143}$ denotes the local terrain height samples, while $\bm{m}_t \in \mathbb{R}^{52}$ indicates the motion command. 

\subsubsection{Action Space}
To ensure knowledge transfer during distillation, all policies share a consistent action space $\mathcal{A} \in \mathbb{R}^{23}$ and employ the same Proportional Derivative (PD) parameters to map the action to joint target positions: $\bm q^{\text{target}}_t = \bm q^{\text{default}} + \bm \alpha \cdot \bm a_t$, where $\bm q^{\text{default}}$ represents the pre-defined default pose and $\bm \alpha$ is a scaling factor following \citet{hm:liao2025beyondmimic}. The joint torques $\bm \tau_t$ are then computed as:
\begin{equation}
    \bm \tau_t = K_p (\bm q^{\text{target}}_t - \bm q_t) - K_d \bm{\dot{q}}_t,
\end{equation}
where $K_p$ and $K_d$ denote the stiffness and damping coefficients, respectively.

\section{Method}
In this section, we introduce the pipeline of the X-Loco framework as shown in Fig~\ref{fig:method}. We first train three specialist policies: upright locomotion $\pi_{l}$, fall recovery $\pi_{r}$, and whole-body coordination $\pi_{w}$. Subsequently, we propose a synergetic policy distillation to effectively consolidate these specialized capabilities into a single generalist policy $\pi_g$.

\subsection{Specialist Policies Training}
All specialists share a unified action space and consistent DoFs. While these specialists are undeployable in real world due to their dependency on privileged state, they achieve peak performance in their respective domains. 

\subsubsection{Upright Locomotion Specialist} 
The upright locomotion specialist $\pi_{l}$ is designed to establish the robot's basic mobility, enabling navigation of diverse terrains, such as stairs, pits, and gaps, while following velocity commands $\bm{c}_t = [v_x, v_y, \omega_z]^\top$. We employ two encoders to process observations, alongside an actor network to predict actions. Specifically, the policy incorporates a history encoder to compress historical states comprising $\{\bm{s}^{\text{prop}}_i, \bm{v}_i, \bm{e}_i, \bm{f}_i\}_{i=t-9}^{t}$ over ten consecutive time steps into a latent representation. Simultaneously, an elevation map encoder processes $\bm{h}_t$ to provide a compact geometric understanding of the surrounding terrain. To ensure behavioral naturalness and enhance exploration efficiency, we incorporate the Adversarial Motion Prior (AMP)~\cite{peng2021amp, tang2024humanmimic} as a style reward to guide the policy. Consequently, $\pi_l$ learns to track velocity commands with a natural gait across diverse terrains, providing upright locomotion guidance for the student policy.

\subsubsection{Fall Recovery Specialist}
The fall recovery specialist $\pi_{r}$ is optimized for robot posture restoration from diverse fallen situations. Given that the objective of $\pi_r$ is postural stabilization rather than velocity-tracking, the policy is designed to be agnostic to terrain information, i.e. $\bm h_{t}$. We adopt an architecture comprising only a history encoder compressing historical state similar as $\pi_l$ and an actor network for $\pi_r$. To ensure the policy can recover from arbitrary failure cases, the robot is initialized in both supine and prone postures during training, with large joint position noise added to the initial states to simulate diverse falling conditions.  Furthermore, an AMP-based style reward is integrated into the training of $\pi_r$ for avoiding jerk motions and guide the policy's exploration. As a result, $\pi_r$ is capable of executing recovery maneuvers, enabling the robot to regain its footing from any fallen state.

\subsubsection{Whole-Body Coordination Specialist}
While existing locomotion policies often lack whole-body coordination ability, we develop the whole-body coordination specialist $\pi_w$ focusing on these skills following a motion tracking paradigm. The $\pi_w$ consists of an elevation map encoder and an actor network. The former processes $\bm h_t$ into a latent representation, which together with $\bm{s}^{\text{prop}}_t$ and $\bm m_t$ is fed into the actor network to predict actions. We employ the elevation map encoder for local terrain perception, thereby enhancing the policy's generalization capability when tracking motions that involve terrain interaction. $\pi_w$ is specialized in tracking motions that necessitate whole-body coordination motion like box climbing and rolling, rather than aiming for universal motion imitation. The subsequent distillation phase transfers these skills into a generalist policy $\pi_g$ and eliminate its dependence on reference motions during inference.

\subsection{Synergetic Policy Distillation}
In the distillation process, we employ a student policy using MoE architecture that maps $\bm s^{\text{prop}}_t$, $\bm c_t$ and $\bm d_t$ to target joint positions, acquiring the capabilities from various specialist policies. To consolidate skills from these specialists into a generalist policy, our synergetic policy distillation employs CASS to dynamically switch specialists based on the robot's state and terrain, while SFI and SAR collectively enhance robustness and training efficiency.

\subsubsection{Case-Adaptive Specialist Selection}
CASS defines three cases according to the robot state and local terrain: recovery, upright locomotion, and coordinated maneuvers. Each case is associated with a specific specialist. Let $\mathcal{C} = \{(c_{\text{rec}}, \pi_r), (c_{\text{loco}}, \pi_l), (c_{\text{coor}}, \pi_w)\}$ denotes the set of all cases and their corresponding specialist. The specific case assigned to the robot is determined by the mapping function $f(b_t, I_t)$, which evaluates the robot's head height $b_t$ and the current terrain context $I_t$ and outputs a specific case index from $\mathcal{C}$. A detailed definition for the mapping function is provided in Appendix C.
During distillation, the target action $a^*_t$ is selected as:
\begin{equation}
a^*_t = \sum_{(i, \pi_i) \in \mathcal{C}} \mathbb{I} (i = f(b_t, I_t)) \cdot \pi_{i}(\bm s_{i,t}),
\end{equation}
where $\bm s_{i,t}$ represents the input of the specialist policy and $\mathbb{I}$ is the indicator function. The termination criteria are consistent with the selected specialist policy to ensure that the collected trajectories remain within state-space manifold explored during specialist training. Notably, each specialist leverages clean privileged observations $\bm s_{i,t}$ for maximum performance, the student policy is trained using a noised non-privileged observation $o_{u,t}$. 
The distillation objective minimizes the mean squared error between the student's action and the selected target action:
\begin{equation}
\mathcal{L}_{\text{distill}} = \mathbb{E} [ \|\pi_g(o_{u,t}) - a^*_t\|_2^2 ].
\end{equation}
Within the case $c_{\text{coor}}$, CASS adaptively modulates the motion commands fed into $\pi_w$, thereby providing the student policy with appropriate whole-body coordination skill guidance across different scenarios. Consequently, CASS provides the student policy with appropriate target action, enabling the generalist humanoid control learning including whole-body coordination, terrain traversal, and fall recovery based on proprioception and depth without requiring reference motions.

\subsubsection{Specialist Annealing Rollout}
Only with the CASS mechanism, training a single policy to master multiple specialist domains simultaneously is challenging. This difficulty stems from two primary issues: 
First, an untrained policy frequently generates trajectories that are out-of-distribution (OOD) relative to the specialists, which are difficult to filter via early termination.
Second, contact-rich and whole-body coordination skills are hard to acquire because errors in the early stages of a maneuver prevent the agent from experiencing subsequent parts of the skill.
To mitigate these challenges, we introduce SAR, which samples actions applied for the environment during data collection by switching between the student and specialist policies according to a mixing ratio $\rho$:
\begin{equation}
a_{env} = b_t a^*_t + (1 - b_t) \pi_g(o_{g,t}),
\end{equation}
where $b_t \sim \text{Bernoulli}(\rho)$ is a binary random variable governed by $\rho$.
This mixed rollout strategy reduces the occurrence of low-quality state manifolds, allowing the student policy to observe and learn from the successful behaviors of complex motion sequences before it can execute them independently. The mixing ratio $\rho$ follows an annealing schedule tied to the training progress. Specifically, we decrease $\rho$ when the distillation loss $\mathcal{L}_{distill}$ falls below a predefined threshold $\epsilon$:
\begin{equation}
\rho \leftarrow \max(0, \rho - \Delta \rho \cdot \mathbb{I}(\mathcal{L}_{distill} < \epsilon))
\end{equation}
By gradually shifting from expert-driven rollouts to autonomous exploration, SAR provides a scaffold for distillation, ensuring the student policy transitions to self-exploration only after attaining a foundational mastery of the specialist’s knowledge.

\subsubsection{Stochastic Fall Injection} 
Instead of solely initialize a subset of environments in fallen states, we propose SFI which introduces random external forces while the robot is walk or climbing, forcing the robot to enter the $c_{rec}$ regime from normal states. To ensure realism, these external force injections are not entirely stochastic, they are conditioned on specific vulnerable scenarios, such as high-speed turns or rolling motions, to simulate unpredictable real-world accidents. To prevent the student policy from being compromised by corrupted data when external forces drive the robot into OOD states relative to $\pi_r$, we implement an additional termination criterion when SFI is triggered. We maintain a buffer of the robot's head height $H$ over consecutive timesteps, the episode is terminated if the variance of the buffer falls below a predefined threshold indicating that the robot has become stuck during the stand-up process:
\begin{equation} 
\mathbb{I} \left( f(b_t, I_t) = c_{rec} \right) \cdot \mathbb{I} \left( \text{Var}(H_{t-k:t}) < \delta \right),
\end{equation}
where $\delta$ is a stability threshold. This criterion ensures that the distillation process remains focused on recoverable states by pruning trajectories where the $\pi_r$ fails to progress.

\begin{figure}[t]
    \centering
    \includegraphics[width=1.0\linewidth]{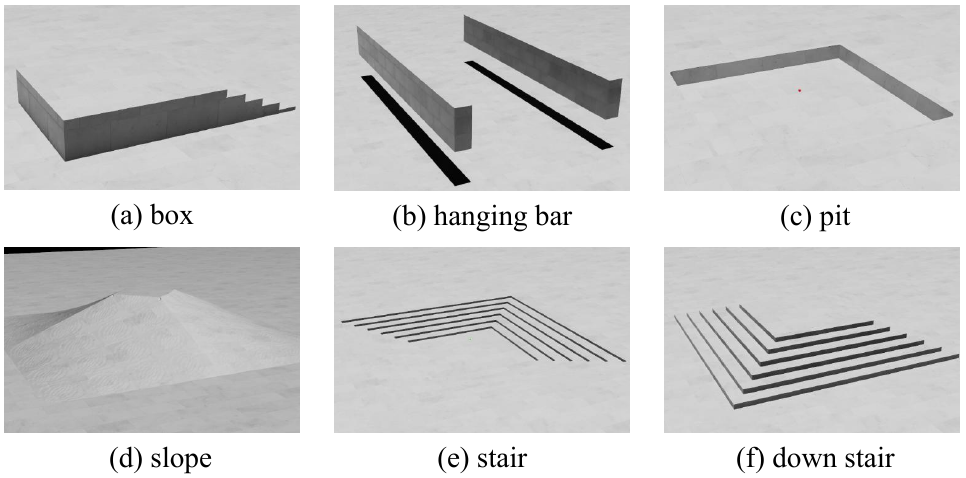}
    \vspace{-0.30cm}
    \caption{Terrains used for training and evaluation.}
    \label{fig:terrains}
\vspace{-0.25cm}
\end{figure}

\subsection{Training Details}
This section details implementation techniques used to optimize the training process. Refer to Appendix A for the complete configurations regarding the simulation environment and hardware setup, as well as reward formulations and domain randomization. 

\subsubsection{Expert Ratio Scheduling}
We implement a hysteresis-based annealing mechanism for the expert ratio $\rho$. Specifically, $\rho$ decays by a step size of $1\mathrm{e}{-4}$ per iteration only when the MSE loss falls below a lower threshold $\tau_{low}=0.005$. Conversely, if the loss exceeds an upper threshold $\tau_{high}=0.010$, the decay process is suspended until the performance recovers (i.e., loss $<\tau_{low}$). This adaptive scheduling dynamically modulates the intensity of expert assistance based on the student's real-time proficiency. The introduction of the hysteresis zone enhances the stability of $\rho$ updates and prevents premature reduction of expert support.

\subsubsection{Stabilizing Whole-Body Specialist Distillation} During the pre-training of $\pi_w$, we incorporate elevation maps into the observation space to facilitate the perception of local terrain geometry. To mitigate overfitting to specific terrain configurations, we randomize the robot's start position with $x$-$y$ offsets sampled from $[-0.15\text{m}, 0.15\text{m}]$ on box terrains. For the distillation phase, we preserve data quality by adopting the specialist's termination criteria to exclude failed samples. 

\subsubsection{Camera Rendering}
Current sensor rendering in simulations~\cite{makoviychuk2021isaac, mittal2025isaac} suffers from low throughput and a lack of visual isolation, where robots inadvertently perceive instances from parallel environments. Furthermore, acceleration frameworks such as NVIDIA Warp~\cite{warp2022} often struggle to concurrently handle both dynamic and static meshes.
We implement a high-performance depth rendering pipeline utilizing NVIDIA Warp to execute parallelized ray-casting. Our approach decouples the environment into static meshes representing the terrain and dynamic meshes representing the robot mesh. To optimize computational throughput, each agent's ray-caster selectively queries the global static mesh and its own local dynamic mesh. 

\section{Experiments} 

\subsection{Experiment Setup}
We perform training and evaluation in the IsaacLab~\cite{mittal2025isaac} simulator. The trained policy is deployed on a Unitree G1 humanoid robot, operating at 50Hz predicting target joint positions. These targets are subsequently tracked by a 500Hz PD controller, which translates them into torques to drive the motors.

In the simulation evaluation, we categorize the test task into three categories: \textbf{Upright Locomotion}, \textbf{Whole-Body Coordination} (\textbf{WBC}), and \textbf{Recovery}. The full suite of evaluation terrains is visualized in Fig. \ref{fig:terrains} and the detailed physical specifications for each terrain type are summarized in Table~\ref{tab:terrain_params}. 

\begin{table*}[t!]
\centering
\caption{Quantitative comparison of X-Loco against baselines and specialist policies. Bold numbers indicates the best performance other than the specialists, and - denotes that the method failed to complete the corresponding task.}
\label{tab:my_results}
\resizebox{2\columnwidth}{!}{
\begin{tabular}{l|cc|cc|cc|c|c|c|c|c} 
\toprule
\multirow{3}{*}{\textbf{Method}} & 
\multicolumn{7}{c|}{\textbf{Locomotion}} & 
\multicolumn{3}{c|}{\textbf{Whole-Body Coordination}} & 
\multicolumn{1}{c}{\textbf{Recovery}} \\
\cmidrule{2-12} 
 & \multicolumn{2}{c|}{Slope} & \multicolumn{2}{c|}{Pit} & \multicolumn{2}{c|}{Stairs} &  & Hanging Bar & Box &  & Flat \\
 & $R_\text{succ}$ & $D_\text{trav}$ & $R_\text{succ}$ & $D_\text{trav}$ & $R_\text{succ}$ & $D_\text{trav}$ & $\bar{R}_\text{succ}$ & $R_\text{succ}$ & $R_\text{succ}$ & $\bar{R}_\text{succ}$ & $R_\text{succ}$ \\
\midrule
BeyondMimic~\cite{hm:liao2025beyondmimic} & - & - & - & - & - & - & - & \textbf{1.000} {\scriptsize $\pm$ .000} & \textbf{0.916} {\scriptsize $\pm$ .008} & \textbf{0.958} {\scriptsize $\pm$ .004} & - \\
MoRE~\cite{hl:wang2025more} & \textbf{0.992} {\scriptsize $\pm$ .008} & 7.863 {\scriptsize $\pm$ .018} & 0.844 {\scriptsize $\pm$ .009} & 6.833 {\scriptsize $\pm$ .114} & 0.926 {\scriptsize $\pm$ .009} & 7.387 {\scriptsize $\pm$ .013} & 0.921 {\scriptsize $\pm$ .009} & - & - & - & - \\
PPO~\cite{schulman2017proximal} & 0.823 {\scriptsize $\pm$ .021} & 6.674 {\scriptsize $\pm$ .152} & 0.793 {\scriptsize $\pm$ .018} & 6.396 {\scriptsize $\pm$ .184} & 0.781 {\scriptsize $\pm$ .025} & 6.565 {\scriptsize $\pm$ .141} & 0.799 {\scriptsize $\pm$ .021} & - & - & - & - \\
AHC~\cite{hl:zhao2025towards} & 0.968 {\scriptsize $\pm$ .005} & 7.871 {\scriptsize $\pm$ .022} & 0.403 {\scriptsize $\pm$ .031} & 3.179 {\scriptsize $\pm$ .254} & 0.278 {\scriptsize $\pm$ .042} & 2.524 {\scriptsize $\pm$ .311} & 0.550 {\scriptsize $\pm$ .026} & - & - & - & \textbf{1.000} {\scriptsize $\pm$ .000} \\
\midrule
Locomotion Specialist & \textcolor{gray}{0.995 {\scriptsize $\pm$ .002}} & \textcolor{gray}{7.989 {\scriptsize $\pm$ .011}} & \textcolor{gray}{0.984 {\scriptsize $\pm$ .010}} & \textcolor{gray}{7.914 {\scriptsize $\pm$ .135}} & \textcolor{gray}{0.991 {\scriptsize $\pm$ .011}} & \textcolor{gray}{7.963 {\scriptsize $\pm$ .004}} & \textcolor{gray}{0.990 {\scriptsize $\pm$ .008}} & - & - & - & - \\
Whole-Body Specialist & - & - & - & - & - & - & - & \textcolor{gray}{1.000 {\scriptsize $\pm$ .000}} & \textcolor{gray}{1.000 {\scriptsize $\pm$ .000}} & \textcolor{gray}{1.000 {\scriptsize $\pm$ .000}} & - \\
Recovery Specialist & - & - & - & - & - & - & - & - & - & - & \textcolor{gray}{1.000 {\scriptsize $\pm$ .000}} \\
\midrule
\textbf{Ours (X-Loco)} & 0.982 {\scriptsize $\pm$ .010} & \textbf{7.984} {\scriptsize $\pm$ .033} & \textbf{0.878} {\scriptsize $\pm$ .015} & \textbf{7.592} {\scriptsize $\pm$ .084} & \textbf{0.958} {\scriptsize $\pm$ .007} & \textbf{7.853} {\scriptsize $\pm$ .011} & \textbf{0.939} {\scriptsize $\pm$ .011} & 0.873 {\scriptsize $\pm$ .014} & 0.868 {\scriptsize $\pm$ .018} & 0.871 {\scriptsize $\pm$ .016} & \textbf{1.000} {\scriptsize $\pm$ .000} \\
\bottomrule
\end{tabular}}
\end{table*}

\begin{table}[t]
\centering
\caption{Terrain Configurations for Evaluation}
\label{tab:terrain_params}
\renewcommand{\arraystretch}{1.3}
\begin{tabular}{llcc}
\toprule
Skill & Obstacle Properties & Ranges & Unit \\ 
\midrule
\multirow{3}{*}{Locomotion} & slope incline      & $[15, 20]$  & $^\circ$ \\
                            & pit obstacle height & $[0.30, 0.40]$   & m        \\
                            & stair step height   & $[0.10, 0.15]$   & m        \\ 
\midrule
\multirow{2}{*}{\begin{tabular}[l]{@{}l@{}}Whole-Body\\ Coordination\end{tabular}} 
                            & obstacle vertical clearance & $[0.87, 0.95]$   & m        \\
                            & obstacle height             & $[0.50, 0.65]$  & m        \\ 
\midrule
Recovery & flat ground & - & - \\
\bottomrule
\end{tabular}
\end{table}

\subsubsection{Baselines}
To validate the locomotion performance of the proposed framework, we compare our framework with the following baselines:
\begin{itemize}
    \item \textbf{BeyondMimic~\cite{hm:liao2025beyondmimic}:} A baseline focuses on whole-body coordination motion tracking relying solely on proprioception. 
    We compare against it to highlight the necessity of exteroception for interacting with environments.
    \item \textbf{MoRE~\cite{hl:wang2025more}:} This baseline employs a two-stage pipeline training a vision-based policy to achieve robust locomotion across complex terrains.
    \item \textbf{AHC~\cite{hl:zhao2025towards}:} This baseline implements an adaptive blind control policy that unifies locomotion with autonomous fall recovery behaviors.
    \item \textbf{PPO~\cite{schulman2017proximal}:} The baseline trains a policy using PPO with depth inputs to traverse across all terrains.
    \item \textbf{Our Specialists:} We evaluate the individual specialist policies ($\pi_l$, $\pi_r$ and $\pi_w$) separately. 
\end{itemize}
To assess the specific contributions of our proposed components, we also evaluate several ablation baselines:
\begin{itemize}
    \item \textbf{Ours w/o CASS:} An ablation to evaluate whether CASS facilitates the acquisition diverse locomotion skills.
    \item \textbf{Ours w/o SFI:} An ablation to evaluate whether SFI helps to enhance the policy's robustness. 
    \item \textbf{Ours w/o SAR:} An ablation to analyze SAR's contribution to the policy distillation process.
    \item \textbf{Ours w/o MoE:} Ablation of policy network with the MoE replaced by a Multi-Layer Perceptron (MLP).
    \item \textbf{MoE-$N$}: Variants of our framework with $N \in \{2, 3\}$ experts to investigate the sensitivity of performance to $N$, where \textbf{MoE-2} serves as our adopted architecture.

\end{itemize}

\subsubsection{Metrics}
We quantify performance using two primary metrics: 
\begin{itemize}
    \item \textbf{Success Rate ($R_\text{succ}$):} The percentage of episodes where the task objective is fully met. Success is defined as: traversing the entire $8\,\text{m}$ track without falling for the \textbf{Upright Locomotion} and \textbf{WBC} task; 
    regaining a standing posture and balancing for $5\,\text{s}$ for the \textbf{Recovery} task. 
    \item \textbf{Traversal Distance ($D_\text{trav}$):} The average distance traveled along the $x$-axis prior to failure or timeout is assessed in the \textbf{Upright Locomotion} tasks.
\end{itemize}

\subsection{Simulation Results}

\subsubsection{Comparative Analysis}
We conduct a comprehensive quantitative comparison between our proposed X-Loco, specialized specialists, and several baselines. The results are summarized in Table \ref{tab:my_results}.

\paragraph{Specialist vs. Baselines} As expected, the specialist policies consistently achieve the highest performance in their respective domains. 
For instance, the locomotion specialist achieves a near-perfect average success rate $\bar R_\text{succ}$ of $0.990$ and the whole-body specialist achieves perfect performance with $\bar R_\text{succ}$ at $1.000$. The superior performance over the baseline in the box climbing task stems from to the utilization of the information in $\bm h_t$.
This results is primarily attributed to the use of privileged information and a focused training objective on a single category of tasks which allows these policies to establish the performance upper bound for each skill.

\paragraph{X-Loco vs. Baselines} Our generalist policy demonstrates significant versatility compared to the baselines. While BeyondMimic excels in \textbf{WBC} tasks with an average $\bar R_\text{succ}$ of $0.958$ but lacks locomotion and recovery capabilities, and MoRE and AHC show competitive results in locomotion but fail completely in \textbf{WBC} task, X-Loco successfully masters all three tasks. In the \textbf{Upright Locomotion} task, X-Loco achieves an average success rate of $0.939$. This performance outperforms the PPO baseline at $0.799$ and AHC at $0.550$ by a large margin, while remaining highly competitive with MoRE. In the \textbf{WBC} task, X-Loco maintains a success rate of $0.871$ without relying on reference motions, whereas other baselines lacking such references fail to execute these tasks.

\paragraph{Specialist vs. X-Loco} The performance gap between X-Loco and the specialist policies is remarkably narrow. In the \textbf{Upright Locomotion} task, X-Loco recovers approximately $94.8\%$ of the specialist's success rate. In the \textbf{Recovery} task, X-Loco matches the specialist with a perfect $1.000$ success rate. While X-Loco exhibits the most pronounced performance degradation in the \textbf{WBC} tasks, this underscores the difficulty of mastering vision-based whole-body coordination skills without reliance on reference motions. These results indicate that X-Loco can effectively distill and integrate diverse expert knowledge into a single policy without significant performance degradation, despite the inherent challenges of multi-task learning and potential gradient interference between different skills.

\begin{figure}[t]
    \centering
    \includegraphics[width=1.0\linewidth]{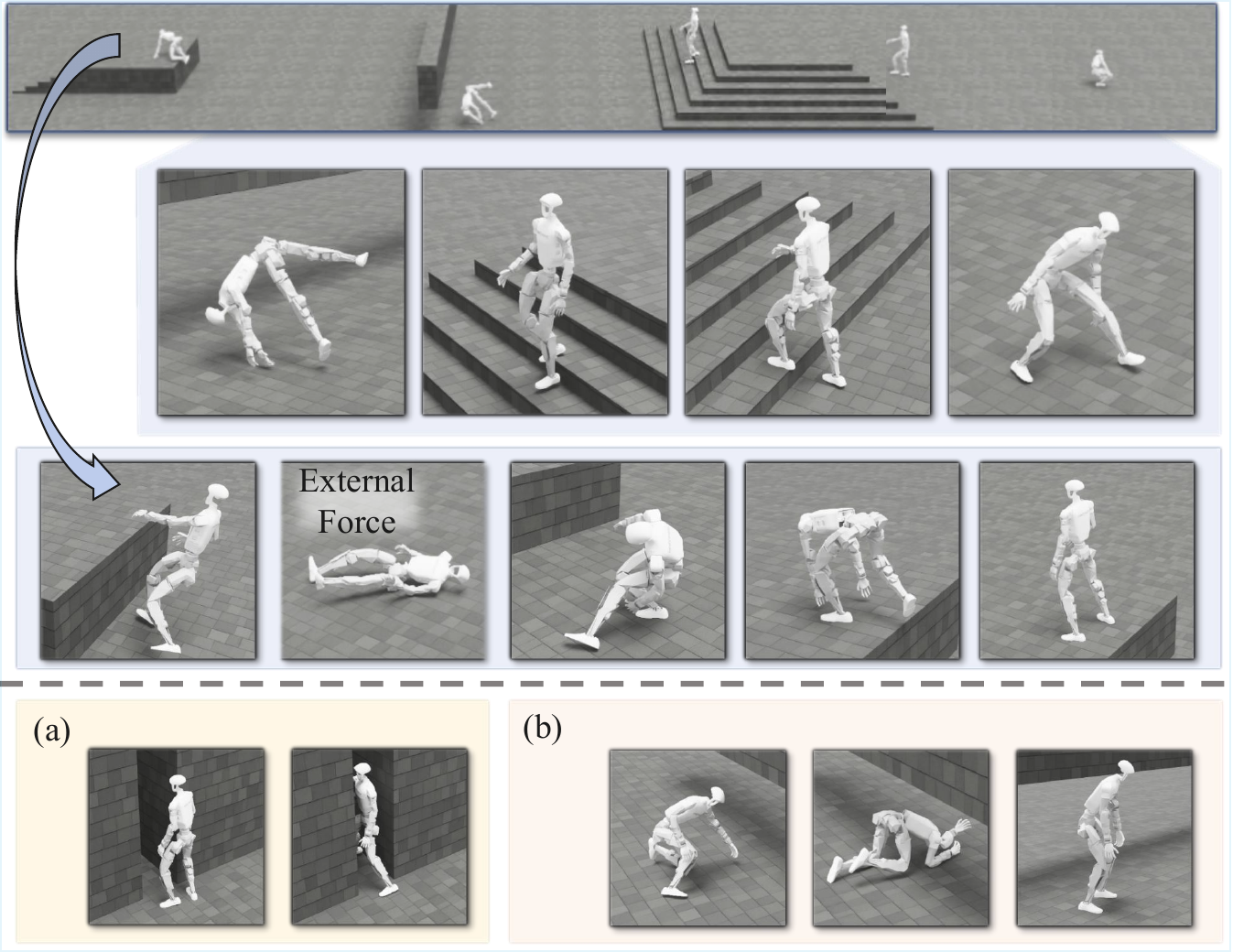}
    \vspace{-0.30cm}
    \caption{\textbf{Top}: Testing the generalist policy on hybrid, challenging terrains. \textbf{Bottom}: Extensibility of the X-Loco to include vision-guided (a) lateral sidling and (b) kneeling crawling.}
    \label{fig:more_analysis}
\vspace{-0.25cm}
\end{figure}

\subsubsection{Ablation Analysis}
We conduct a comprehensive ablation study on CASS, MoE architecture, SAR and SFI. 

\begin{table}[t!]
\centering
\caption{Quantitative results of ablation analysis on CASS and MoE architectures.}
\label{tab:ablation}
\resizebox{1.0\columnwidth}{!}{
\begin{tabular}{l|c|c|c|c} 
\toprule
\multirow{2}{*}{\textbf{Method}} & \textbf{Locomotion} & \textbf{\shortstack{Whole-Body \\ Coordination}} & \textbf{Recovery} & \textbf{Average} \\
 & $\bar{R}_{\text{succ}}$ & $\bar{R}_{\text{succ}}$ & $R_{\text{succ}}$ & $\bar{R}_{\text{succ}}$ \\
\midrule
Ours w/o CASS & 0.903 & 0.446 & \textbf{1.000} & 0.783 \\
Ours w/o MoE & 0.692 & 0.561 & 0.996 & 0.749 \\
MoE-3 & 0.628 & 0.709 & \textbf{1.000} & 0.779 \\
\textbf{MoE-2 (Ours)} & \textbf{0.939} & \textbf{0.845} & \textbf{1.000} & \textbf{0.928} \\
\bottomrule
\end{tabular}}
\end{table}

\paragraph{Effectiveness of CASS}
To evaluate the contribution of CASS, we compare our method against the variant \textbf{Ours w/o CASS}. As reported in Table~\ref{tab:ablation}, the absence of CASS leads to a drastic performance collapse in both \textbf{WBC} and \textbf{Upright Locomotion} tasks. 
The primary driver of this degradation is that without the selection mechanism, the distillation process lacks critical training samples representing the transition phases between heterogeneous motor patterns. Specifically, the student policy is not exposed to the specific state-action pairs required to bridge disparate skills. 
In summary, CASS is essential for providing the guidance to synthesize specialized specialist into a generalist policy, ensuring the emergent generalist policy maintains high fidelity to specialist capabilities while mastering the connective dynamics between them.

\paragraph{Ablations on MoE Architecture}
We evaluate \textbf{Our w/o MoE} and \textbf{MoE-3} across all test tasks and report the average success rate. As shown in Table~\ref{tab:ablation}, the \textbf{Ours w/o MoE} variant struggles to maintain high success rates across all tasks, with performance dropping significantly in \textbf{WBC} tasks. This confirms that the MLP backbone lacks the capacity to reconcile state-action distributions from multiple specialists into a shared parameter space. 
While the \textbf{MoE-2} significantly recovers performance, further increasing the number of experts (\textbf{MoE-3}) leads to a regression, particularly in \textbf{Upright Locomotion} tasks. 
We attribute this to the fact that an excessive number of experts can lead to sparse updates and sub-optimal expert utilization during the distillation process.

\begin{figure}[t]
  \centering
  \begin{minipage}[t]{0.48\linewidth}
    \vspace{0pt}
    \centering
    \includegraphics[width=1.0\linewidth]{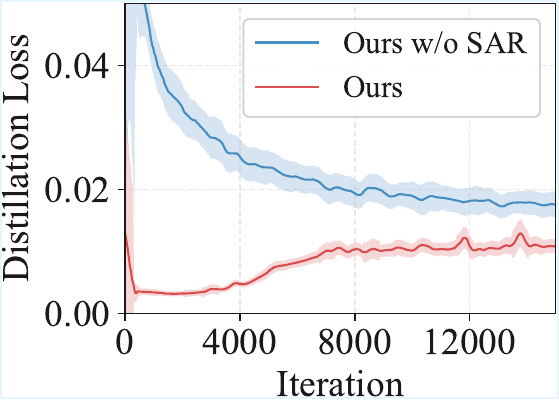}
    \vspace{-0.2cm}
    \captionof{figure}{Distillation loss curves of ablation on SAR.}
    \label{fig:sar_result}
  \end{minipage}
  \hfill
  \begin{minipage}[t]{0.48\linewidth}
    \vspace{10pt}
    \centering
    \resizebox{0.8\linewidth}{!}{
      \begin{tabular}{lcc}
        \toprule
        \textbf{Method} & \begin{tabular}[c]{@{}c@{}} \textbf{Recovery} \\ $R_{\text{succ}}$ \end{tabular}  \\
        \midrule
        Ours w/o SFI  & 0.912  \\
        \textbf{Ours}  & \textbf{0.958}  \\
        \bottomrule
      \end{tabular}
    }
    \vspace{0.6cm}
    \captionof{table}{Recovery success rate of ablation on SFI with large external disturbance}
    \label{tab:sfi_results}
  \end{minipage}
\end{figure}

\paragraph{Ablations on SAR \& SFI}
As shown in Fig.~\ref{fig:sar_result}, our method demonstrates a clear advantage in distillation loss over the \textbf{Ours w/o SAR} baseline. In the early training phase, the loss decreases rapidly as the policy focuses on high-quality samples from the specialists. As $\rho$ decays, the student policy begins to utilize its own rollout data including non-optimal or failed trajectories to learn to adapt complex situations, leading to a rise in loss. Ultimately, X-Loco achieves faster convergence to a lower terminal loss. This confirms that SAR effectively bridges specialist guidance and self-exploration, ensuring more efficient policy distillation. To evaluate the role of SFI, we assess the policy’s performance in \textbf{Upright Locomotion} and \textbf{WBC} tasks under large external disturbances that lead to falls. We measure the success rate after falling to quantify the recovery capability. As shown in Table~\ref{tab:sfi_results}, the variant with SFI achieves a higher success rate, demonstrating that SFI significantly enhances the policy's ability to regain balance in unpredictable environments.

\begin{figure*}[t]
    \centering
    \includegraphics[width=1.0\linewidth]{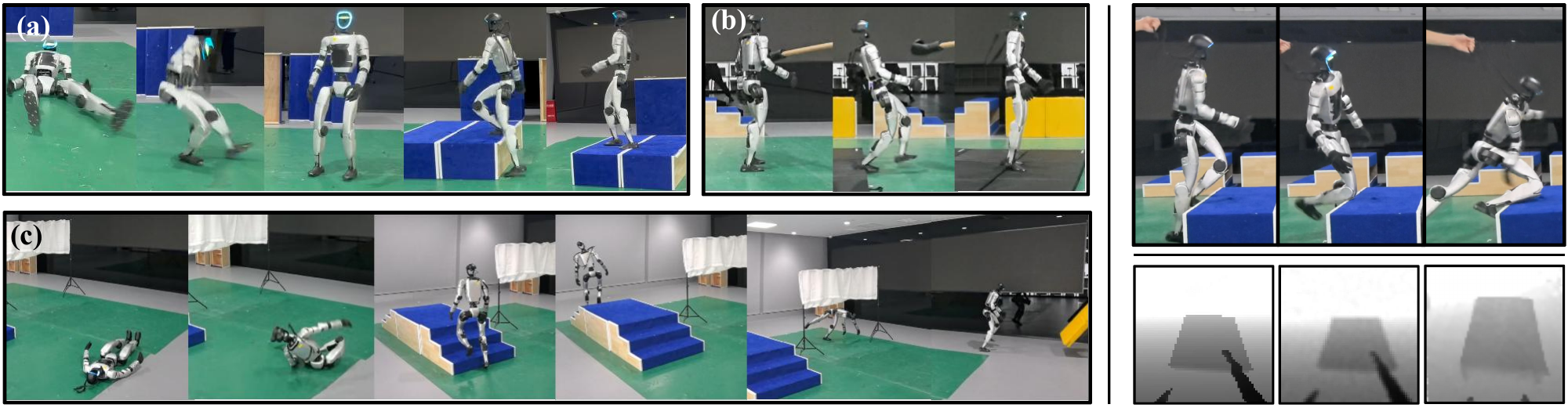}
    \caption{\textbf{Left}: (a) fall recovery and platform traversal; (b) resilience to external disturbances; (c) a continuous sequence of recovery, stair climbing, and rolling under an overhead bar. \textbf{Right}: (Top) a failure case where the robot trips due to lack of camera randomization. (Bottom, from left to right) original depth in simulation, noisy depth, and processed real-world depth.}
    \label{fig:deploy}
\vspace{-0.25cm}
\end{figure*}

\subsection{More Analysis}
\subsubsection{Hybrid Terrain}
We further evaluated the robustness and skill integration of the generalist policy by constructing a challenging hybrid terrain sequence that requires continuous navigation across diverse obstacles. In this experiment, the robot is initialized in a fallen state and must sequentially perform fall recovery, climb up and down stairs, roll under a hanging bar, and finally climb a box. This is a long-horizon task using depth information, as shown in Fig~\ref{fig:more_analysis}. It can even recover after being pushed off a high platform and subsequently resume the climbing task. This successful traversal across such a challenging terrain proves that our framework effectively consolidates diverse skills into a single policy, enabling the robot to execute multifaceted movements while maintaining inherent adaptability.

\subsubsection{Framework Extensibility}
X-Loco enables the capability expansion of the generalist policy by expanding motions learned by the whole-body coordination specialist and subsequently adjusting CASS and terrains during the distillation process.  Fig.~\ref{fig:more_analysis} illustrates two examples of capability expansion: lateral movement through narrow gaps and crawling under low-overhanging obstacles. The results demonstrate that the distilled generalist policy $\pi_g$ successfully mastered these skills: when encountering narrow gaps, the robot autonomously transitions to a sidling posture to traverse the terrain; when faced with low-clearance obstacles, the robot spontaneously adopts a crawling gait to pass under the obstacles. This capability underscores the versatility and extensibility of our framework. By simply enriching the specialist's training set, the generalist policy can effectively integrate new whole-body coordination skills, proving that X-Loco is a scalable solution for humanoid locomotion control.

\subsection{Real-World Deployment}
\paragraph{Performance on Hybird Terrains}
To validate the robustness and versatility of X-Loco in the real world, we deployed the generalist policy on a Unitree G1 robot. Fig.~\ref{fig:cover} illustrates the performance of X-Loco across diverse challenging terrains, including a 90cm-high hanging bar and a 60cm-tall box. In a series of comprehensive tests, the robot successfully manifested a seamless transition between the three core locomotion abilities. Fig.~\ref{fig:deploy} demonstrates X-Loco’s performance across various composite terrains under disturbances. The policy successfully executes a series of continuous maneuvers, including recovering from a fall, traversing high platforms, climbing stairs, and performing a rolling motion to pass under a hanging bar. Furthermore, the experimental results confirm the policy's exceptional resilience. When subjected to severe external disturbances or manual pushes that result in a fall, the policy autonomously regains an upright posture and resumes locomotion. The experiment results demonstrate that the capabilities integrated via our distillation pipeline generalize effectively to physical hardware.

\paragraph{Depth Sim-to-Real Analysis}
Transitioning the policy from simulation to the real world reveals a large depth sim-to-real gap, which significantly degrades the policy's performance as shown in Fig.~\ref{fig:deploy}. The policy suffers from perception-action mismatches caused by depth disparities, resulting in collisions with obstacles followed by falls or continuous foot scuffing against the ground during locomotion. To bridge the depth sim-to-real gap, we implement a multi-stage pipeline that synthesizes realistic depth data by injecting additive Gaussian noise and blurring to emulate real-world sensor. Furthermore, we employ domain randomization over the camera's extrinsic and intrinsic parameters, including random perturbations of its orientation, position, and field of view (FOV). For real-world depth images, we apply hole-filling and Gaussian filtering to mitigate sensor noise and data sparsity. As illustrated in Fig.~\ref{fig:deploy}, this preprocessing ensures consistent image alignment between simulation and reality, enabling the policy to achieve successful sim-to-real transfer.

\section{Conclusion} 
We present X-Loco, a framework that enables a single vision-based policy to achieve generalist humanoid control via velocity commands. By employing synergetic policy distillation, X-Loco consolidates three specialist policies into a unified agent. SAR and SFI are introduced to bridge the gap between expert guidance and autonomous exploration, and to enhance the policy’s recovery resilience, respectively. Experimental results demonstrate that X-Loco achieves successful integration of diverse locomotion skills with performance comparable to that of the specialists, offering a scalable solution for generalist humanoid locomotion. Real-world deployments further confirm its capability to manage complex environments, pushing the boundaries of humanoid locomotion.

\section{Limitations and Future Work}
Despite its advancements, X-Loco has limitations. Its performance is primarily constrained by a limited sensory horizon due to the narrow-FOV camera, and the difficulty of perfectly modeling real-world sensor noise. Additionally, since the distillation relies on behavior cloning, the policy is inherently bounded by the specialists’ performance, making X-Loco less effective in edge cases lacking expert coverage.

Future work will follow two directions. First, we will integrate multi-modal sensing (e.g., RGB-D and LiDAR) to broaden the perceptual field and rectify misjudgments through cross-modal calibration. Second, we aim to develop a hybrid learning framework that combines distillation with RL-based fine-tuning, allowing the policy to explore and generalize beyond specialist demonstrations.

\bibliographystyle{plainnat}
\bibliography{references}

\clearpage

\appendix

\subsection{Experimental Setup Details}
\subsubsection{Simulation Environment and Robot Configuration}
We conduct training and evaluation in the Isaac Lab simulator~\cite{mittal2025isaac}. The physics simulation operates at $200$ Hz, while the policy runs at $50$ Hz, with a decimation factor of $4$. For real-world evaluation, we use the Unitree G1 humanoid robot, which is equipped with an Intel RealSense D435i and has 23 actuated degrees of freedom (6 per leg, 4 per arm, and 3 in the waist). The robot is equipped with a Jetson Orin NX for onboard computation. We assign different $K_p, K_d$ parameters and action scales for each joint as shown in Table~\ref{tab:pd_gains}, following the setup in~\cite{hm:liao2025beyondmimic}.

\begin{table}[h]
\centering
\caption{PD Controller Parameters and Action Scaling}
\label{tab:pd_gains}
\renewcommand{\arraystretch}{1.2}
\resizebox{\linewidth}{!}{
\begin{tabular}{lccc}
\toprule
Joint Names & $K_p$ (N$\cdot$m/rad) & $K_d$ (N$\cdot$m$\cdot$s/rad) & Action Scale \\
\midrule
Hip Roll, Knee & $99.10$ & $6.31$ & $0.35$ \\
Hip Pitch, Hip Yaw, Waist Yaw & $40.18$ & $2.56$ & $0.55$ \\
Ankle (P/R), Waist (R/P) & $28.50$ & $1.81$ & $0.44$ \\
Shoulder (P/R/Y), Elbow & $14.25$ & $0.91$ & $0.44$ \\
\bottomrule
\end{tabular}
}
\end{table}

\subsubsection{Reward Formulation}
The formulation of reward functions and their respective weights for the three specialist policies ($\pi_\text{l}, \pi_\text{r}, \pi_\text{w}$) are detailed in Table~\ref{tab:reward_pi_w}--\ref{tab:reward_pi_r}. To facilitate the emergence of naturalistic motion patterns, we incorporate an Adversarial Motion Prior (AMP) reward into the training of the upright locomotion ($\pi_\text{l}$) and fall recovery ($\pi_\text{r}$) specialists. The mathematical formalization of the AMP reward and its optimization objectives are provided in Appendix B.

\subsubsection{Domain Randomization}
We implement a consistent Domain Randomization (DR) configuration during the specialist policy training and generalist policy distillation to facilitate robust sim-to-real transfer. Our DR mainly involves randomizing physical parameters, initial states, and actuator dynamics. Specifically, we vary ground friction and inertial properties to handle environmental diversity, while applying external force perturbations to simulate real-world uncertainties. The full randomization configuration is listed in Table~\ref{tab:domain_randomization}.

\begin{table}[h]
\centering
\caption{Domain Randomization Parameters. ($\mathcal{U}[\cdot]$: uniform distribution)}
\label{tab:domain_randomization}
\renewcommand{\arraystretch}{1.3}
\resizebox{\linewidth}{!}{
\begin{tabular}{lcc}
\toprule
Domain Randomization & Sampling Distribution & Unit \\ 
\midrule
Static friction & $\mu_{\text{static}} \sim \mathcal{U}[0.6, 1.0]$ & - \\
Dynamic friction & $\mu_{\text{dynamic}} \sim \mathcal{U}[0.4, 0.8]$ & - \\
Torso payload mass & $m_{\text{payload}} \sim \mathcal{U}[-1.0, 5.0]$ & kg \\ 
Initial joint positions & $\theta_{\text{init}} \sim \mathcal{U}[0.8, 1.2] \times \theta_{\text{nominal}}$ & - \\
External push interval & $\Delta t_{\text{push}} \sim \mathcal{U}[5.0, 8.0]$ & s \\
Push linear velocity & $v_x,v_y \sim \mathcal{U}[-0.5, 0.5], v_z \sim \mathcal{U}[-0.2, 0.2]$ & m/s \\
Push angular velocity & $\omega_r, \omega_p \sim \mathcal{U}[-0.52, 0.52], \omega_y \sim \mathcal{U}[-0.78, 0.78]$ & rad/s \\
Actuator stiffness $K_p$ & $k_p \sim \mathcal{U}[0.8, 1.2] \times K_{p,\text{nominal}}$ & - \\
Actuator damping $K_d$ & $k_d \sim \mathcal{U}[0.8, 1.2] \times K_{d,\text{nominal}}$ & - \\
\bottomrule
\end{tabular}
}
\end{table}

\subsubsection{Terrain Curriculum}
We adopt an adaptive terrain curriculum inspired by~\cite{rudin2022learning} for the training of $\pi_l$ and $\pi_g$, which automatically modulates terrain difficulty based on agent performance. The environment is structured as a $10 \times 20$ grid of $8\text{m} \times 8\text{m}$ patches, encompassing five distinct terrain types: slopes, pits, hanging bars, stairs, and boxes. The difficulty is scaled by adjusting geometric parameters: slopes range from $0^{\circ}$ to $20^{\circ}$; pits vary in height from $0.05\text{m}$ to $0.3\text{m}$; hanging bars enforce vertical clearance constraints of $0.67\text{m}$ to $0.72\text{m}$; stair step heights range from $0\text{m}$ to $0.15\text{m}$; and climbing boxes vary in height from $0.45\text{m}$ to $0.65\text{m}$.

\subsubsection{Training Implementation Details}
In this section, we detail the neural network architectures and the training parameters for both the specialist and generalist policies.

\paragraph{Specialist Policy Architecture}
The specialist policies ($\pi_\text{l}, \pi_\text{r}, \pi_\text{w}$) are trained using the Actor-Critic framework and comprise some or all of the following modules:
\begin{itemize}
    \item \textbf{History Encoder:} An MLP with hidden units $[1024, 512, 128]$, utilized by $\pi_\text{l}, \pi_\text{r}$.
    \item \textbf{Elevation Map Encoder:} A 3-layer MLP $[128, 64, 32]$ employed by $\pi_\text{l}$ and $\pi_\text{w}$ to process local terrain geometry.
    \item \textbf{Privileged Proprioception Encoder:} A 3-layer MLP $[128, 64, 32]$ for $\pi_\text{l}$ to encode privileged states.
    \item \textbf{Actor Head and Critic:} Both networks are modeled as 3-layer MLPs with dimensions $[512, 256, 128]$. For the fall recovery specialist $\pi_\text{r}$, we adopt a multi-critic structure~\cite{hr:huang2025learning} comprising four independent critics.
\end{itemize}

\paragraph{Generalist Policy Architecture}
The generalist policy $\pi_g$ leverages a Mixture-of-Experts (MoE) architecture, where a gating network modulates the contributions of experts. The experts and the gating network are instantiated as 3-layer MLPs $[512, 256, 128]$. To incorporate exteroception, a CNN-based depth encoder is employed to compress sequential depth images into a 128-dimensional latent representation. In parallel, a history encoder is utilized to process proprioceptive sequences over a 10-step horizon. This temporal context, combined with the latent representation from the depth encoder, is subsequently fed into the MoE to predict the actions.

\paragraph{Training Hyperparameters}
Specialist policies are trained via PPO~\cite{schulman2017proximal}, while the generalist policy is distilled using supervised behavior cloning.
Detailed training hyperparameters are provided in Table~\ref{tab:ppo_hyperparams}.

\begin{table}[h]
\centering
\caption{Optimization Hyperparameters for Specialists and Generalist Policies}
\label{tab:ppo_hyperparams}
\resizebox{\linewidth}{!}{
\renewcommand{\arraystretch}{1.2}
\small
\begin{tabular}{lcc}
\toprule
\textbf{Hyperparameter} & \textbf{Specialist Training (PPO)} & \textbf{Generalist Distillation} \\
\midrule
Number of environments & 4096 & 4096 \\
Learning rate & $1.0 \times 10^{-3}$ (Adaptive) & $1.0 \times 10^{-3}$ (Fixed) \\
Num. epochs per iteration & 5 & 8 \\
Num. mini-batches & 4 & 12 \\
Steps per training batch & 24 & 12\\
Discount factor ($\gamma$) & 0.99 & - \\
GAE parameter ($\lambda$) & 0.95 & - \\
PPO clip parameter ($\epsilon$) & 0.2 & - \\
Entropy coefficient & 0.005 & - \\
Desired KL divergence & 0.01 & - \\
\midrule
\textbf{Total Training Iterations} & \multicolumn{2}{c}{} \\
\midrule
Upright Locomotion ($\pi_\text{l}$) & \multicolumn{2}{c}{30,000} \\
Recovery ($\pi_\text{r}$) & \multicolumn{2}{c}{10,000} \\
Whole-Body Coordination ($\pi_\text{w}$) & \multicolumn{2}{c}{50,000} \\
Generalist Policy ($\pi_\text{g}$) & \multicolumn{2}{c}{30,000} \\
\bottomrule
\end{tabular}
}
\end{table}

\subsubsection{Training Pseudocode}
We present the training procedure for the generalist policy $\pi_\text{g}$ in Algorithm~\ref{alg:synergetic_distillation}.
\begin{algorithm}[t]
\caption{Synergetic Policy Distillation}
\label{alg:synergetic_distillation}
\begin{algorithmic}[1]
\REQUIRE Set $\mathcal{C} = \{(c_{rec}, \pi_r), (c_{loco}, \pi_l), (c_{coor}, \pi_w)\}$, ratio $\rho$, decay $\Delta\rho$, threshold $\epsilon$
\STATE Initialize storage buffer $\mathcal{D}$ and student policy $\pi_{\theta}$
\FOR{iteration = $1, 2, \dots$}
    \STATE Clear storage $\mathcal{D}$
    \STATE // \textbf{Data Collection with SAR and SFI}
    \FOR{step $t = 1$ to $T$ in $N$ parallel environments}
        \STATE Get student state $o_{u,t}$ and privileged states $s_{i,t}$.
        \STATE \textbf{SFI:} Inject external forces based on context.
        \STATE Determine case index $i \gets f(b_t, I_t)$ and select corresponding specialist $\pi_i$.
        \STATE Get expert action: $a^*_t \gets \pi_i(s_{i,t})$.
        \STATE Get student action: $a_t^u \gets \pi_u(o_{u,t})$.
        
        \STATE \textbf{SAR:} Sample $b_t \sim \text{Bernoulli}(\rho)$:
        \STATE $a_{env} \gets b_t a^*_t + (1 - b_t) a_t^u$
        
        \STATE Execute $a_{env}$, and store $(o_{u,t}, a^*_t)$ in $\mathcal{D}$.
        \STATE \textbf{Termination:} \textbf{if} specialist $\pi_i$ termination criteria met \textbf{or} SFI-specific failure (Eq. 5) \textbf{then} reset env
    \ENDFOR
    
    \STATE // \textbf{Policy Optimization}
    \FOR{epoch = $1$ to $K$}
        \STATE Sample minibatches from $\mathcal{D}$.
        \STATE $\mathcal{L}_{distill} \gets \text{MSE}(\pi_u(o_{u,t}), a^*_t)$.
        \STATE Update $\theta$ via gradient descent to minimize $\mathcal{L}_{distill}$.
    \ENDFOR
    
    \STATE // \textbf{Specialist Annealing Schedule}
    \IF{$\mathcal{L}_{distill} < \epsilon$}
        \STATE $\rho \gets \max(0, \rho - \Delta\rho)$
    \ENDIF
\ENDFOR
\end{algorithmic}
\end{algorithm}

\begin{table}[h]
\centering
\caption{Depth Augmentation and Camera Randomization Parameters. ($\mathcal{U}[\cdot]$: uniform distribution)}
\label{tab:depth_augmentation}
\renewcommand{\arraystretch}{1.3}
\resizebox{\linewidth}{!}{
\begin{tabular}{lcc}
\toprule
\textbf{Camera Randomization} & \textbf{Value / Range} & \textbf{Unit} \\
\midrule
Gaussian noise & $\sigma_{\text{noise}} = 0.02$ & m \\
Gaussian filter kernel & $k \in \{3 \times 3\}$ & pixel \\
Gaussian filter sigma & $\sigma_{\text{filter}} = 1.0$ & - \\
Camera position & $\Delta x, \Delta y, \Delta z \sim \mathcal{U}[-0.05, 0.05]$ & m \\
Camera rotation & $\theta_{\text{pitch}} \sim \mathcal{U}[-10, 5]$, $\theta_{\text{roll}}, \theta_{\text{yaw}} \sim \mathcal{U}[-1, 1]$ & $^\circ$ \\
Camera horizontal FOV & $\Delta \text{FOV}_{\text{h}} \sim \mathcal{U}[-10, 10]$ & $^\circ$ \\
\bottomrule
\end{tabular}
}
\end{table}

\subsection{Adversarial Motion Prior (AMP) Formulation}
To facilitate naturalistic motion patterns, we incorporate the Adversarial Motion Prior (AMP) framework \cite{peng2021amp, escontrela2022adversarial} into the training of the locomotion ($\pi_\text{l}$) and recovery ($\pi_\text{r}$) specialists. The AMP module produces a style-dependent reward $r_t^{\text{style}}$, by training a discriminator $D_{\phi}$ to differentiate between reference trajectories and policy-generated ones. 

The discriminator operates on a compact input representation $\tau_t = (s_{t-3:t+1}^{\text{amp}})$, which comprises a 5-step sequence of AMP states. Each individual AMP state $s_t^{\text{amp}} \in \mathbb{R}^{23}$ consists of the robot's joint positions. Following the Least Squares GAN (LSGAN) \cite{mao2017least} formulation, the discriminator is optimized by maximizing the following objective function:
\begin{equation}
\begin{split}
\mathcal{L}_D = &\ \mathbb{E}_{\tau \sim \mathcal{M}}[(D_{\phi}(\tau) - 1)^2] + \mathbb{E}_{\tau \sim \mathcal{P}}[(D_{\phi}(\tau) + 1)^2] \\
& + \frac{\alpha^{d}}{2} \mathbb{E}_{\tau \sim \mathcal{M}}[\|\nabla_{\phi} D_{\phi}(\tau)\|_2^2],
\end{split}
\label{eq:disc_loss}
\end{equation}
where $\mathcal{M}$ and $\mathcal{P}$ denote the reference dataset and the on-policy rollout buffer, respectively. Training stability is maintained via a gradient penalty with coefficient $\alpha^d$ applied to the reference data. 

The discriminator's output $d = D_{\phi}(\tau_t)$ is subsequently mapped to a surrogate reward $r_t^{\text{style}}$ to guide the policy:
\begin{equation}
    r^{\text{style}}(s_t) = w_{\text{style}} \cdot \max\left(0,\, 1 - \frac{1}{4}(d - 1)^2\right),
\end{equation}
where $w_{\text{style}}$ is a scaling coefficient. The composite reward used for policy optimization is the summation of task-specific and style-dependent components: $r_t = r_t^{\text{task}} + r_t^{\text{style}}$.

Reference motions for the locomotion specialist ($\pi_\text{l}$) are source from the LAFAN1 dataset \cite{harvey2020robust}, retargeted by Unitree. For the recovery specialist ($\pi_\text{r}$), reference motions are sourced from motion capture (MoCap) data and retargeted to the Unitree G1 platform.

\subsection{Case-Adaptive Specialist Selection Details}
\begin{figure}[h]
    \centering
    \includegraphics[width=1.0\linewidth]{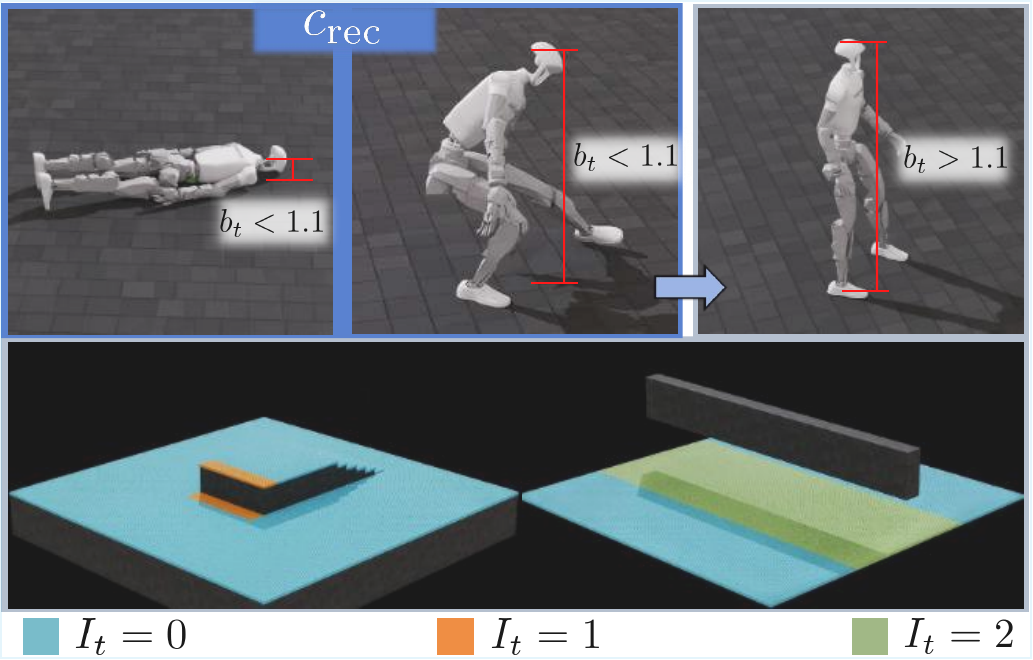}
    \vspace{-0.30cm}
    \caption{Illustration of the case-adaptive specialist selection process based on $b_t$ and $I_t$.}
    \label{fig:cass}
\vspace{-0.25cm}
\end{figure}

The Case-Adaptive Specialist Selection (CASS) mechanism selects an appropriate specialist to guide the student policy based on the robot's head height $b_t$ and the environmental context $I_t$. This mechanism incorporates a mapping function $f(b_t, I_t)$ that identifies the robot's specific case, alongside a set $\mathcal{C} = \{(c_{\text{rec}}, \pi_r), (c_{\text{loco}}, \pi_l), (c_{\text{coor}}, \pi_w)\}$ that maps these cases to their respective specialists. Specifically, if the robot's head height $b_t$ is below $1.1m$, it is categorized as case $c_{\text{rec}}$, where the fall recovery policy $\pi_r$ provides guidance. If $b_t$ exceeds $1.1m$, the classification depends on the local environmental context $I_t$. In scenarios where the robot is in proximity to a hanging bar or a high platform, it is assigned to case $c_{\text{coor}}$ and guided by $\pi_w$; otherwise, it is classified as $c_{\text{loco}}$ and supervised by $\pi_l$. Once $\pi_w$ is selected, CASS provides reference motions tailored to different terrains, enabling $\pi_w$ to generate the appropriate guidance actions. Consequently, $c_{\text{coor}}$ includes into several sub-cases $c_{\text{coor}}^i$ based on the specific reference motion requirements. The definition of $I_t$ is illustrated in Figure~\ref{fig:cass}. Accordingly, $f(b_t, I_t)$ can be expressed as:

\begin{equation}
f(b_t, I_t) = 
\begin{cases} 
c_{\text{rec}}, & \text{if } b_t < 1.1 \\
c_{\text{coor}}^1, & \text{if } b_t \geq 1.1 \text{ and } I_t =1 \\
c_{\text{coor}}^2, & \text{if } b_t \geq 1.1 \text{ and } I_t =2 \\
c_{\text{loco}}, & \text{otherwise}
\end{cases}, 
\end{equation}
where $I_t=1$ signifies that the robot is facing a high platform requiring a climbing maneuver, while $I_t=2$ represents the presence of a hanging obstacle approximately $0.7m$ above the ground that can be navigated by forward rolling.

\vspace{-0.2cm}

\subsection{Whole-Body Coordination Specialists $\pi_\text{w}$ Details}
To facilitate local terrain awareness, the whole-body coordination specialists $\pi_{w}$ utilizes a terrain height-scanner centered on the robot's pelvis to get the local terrain height samples (i.e., an elevation map). The sensor captures the local terrain geometry as a grid of $1.6m\times1.0m$ with a spatial resolution of $0.1m$. Furthermore, we employ a random initialization strategy. In Box terrain, the robot is initialized at a longitudinal distance sampled from $\mathcal{U}[0, 0.55]m$ ralative to the box. For Hanging Bar terrain, the robot is initialized at a distance sampled from $\mathcal{U}[3, 3.5]m$ from the bar. This initialization strategy ensures that $\pi_\text{w}$ encounters a diverse set of relative distances to obstacles, thereby improving the ability to handle environmental variability.

\subsection{Depth Alignment for Sim-to-Real Transfer}
To bridge the visual sim-to-real gap, we implement a process that matches simulated depth distributions to real-world sensor characteristics. By injecting Gaussian noise, blur, and randomizing camera intrinsic and extrinsic parameters, we approximate the sensor observations of the actual hardware. The specific noise and randomization parameter ranges are detailed in Table~\ref{tab:depth_augmentation}.

\begin{table}[ht]
\centering
\caption{Reward Function Definitions for the Whole-Body Coordination Specialist ($\pi_w$).}
\label{tab:reward_pi_w}
\resizebox{\linewidth}{!}{
\renewcommand{\arraystretch}{1.4}
\begin{tabular}{llc}
\toprule
\textbf{Reward Terms} & \textbf{Equation} & \textbf{Weight} \\
\midrule
Body position & $\exp\left( - \left( \frac{1}{|\mathcal{B}_{\text{target}}|} \sum_{b \in \mathcal{B}_{\text{target}}} \|\mathbf{p}_b^{\text{des}} - \mathbf{p}_b\|^2 \right) / 0.3^2 \right)$ & 1.0 \\
\addlinespace
Body orientation & $\exp\left( - \left( \frac{1}{|\mathcal{B}_{\text{target}}|} \sum_{b \in \mathcal{B}_{\text{target}}} \|\log(R_b^{\text{des}} R_b^\top)\|^2 \right) / 0.4^2 \right)$ & 1.0 \\
\addlinespace
Body linear velocity & $\exp\left( - \left( \frac{1}{|\mathcal{B}_{\text{target}}|} \sum_{b \in \mathcal{B}_{\text{target}}} \|\mathbf{v}_b^{\text{des}} - \mathbf{v}_b\|^2 \right) / 1.0^2 \right)$ & 1.0 \\
\addlinespace
Body angular velocity & $\exp\left( - \left( \frac{1}{|\mathcal{B}_{\text{target}}|} \sum_{b \in \mathcal{B}_{\text{target}}} \|\boldsymbol{\omega}_b^{\text{des}} - \boldsymbol{\omega}_b\|^2 \right) / 3.14^2 \right)$ & 1.0 \\
\addlinespace
Anchor position & $\exp\left( - \|\mathbf{p}_{\text{anchor}}^{\text{des}} - \mathbf{p}_{\text{anchor}}\|^2 / 0.3^2 \right)$ & 0.5 \\
\addlinespace
Anchor orientation & $\exp\left( - \|\log(R_{\text{anchor}}^{\text{des}} R_{\text{anchor}}^\top)\|^2 / 0.4^2 \right)$ & 0.5 \\
Action smoothness & $\|\mathbf{a}_t - \mathbf{a}_{t-1}\|^2$ & $-0.1$ \\
\addlinespace
Joint position limit & $\sum_{j=1}^N \left[ \max(l_j - \theta_j, 0) + \max(\theta_j - u_j, 0) \right]$ & $-10.0$ \\
\addlinespace
Undesired self-contacts & $\sum_{b \notin \mathcal{B}_{\text{ee}}} \mathbf{1} [ \|f_b^{\text{self}}\| > 1\text{N} ]$ & $-0.1$ \\
\bottomrule
\end{tabular}
}
\end{table}

\begin{table}[h]
\centering
\caption{Reward Function Definitions for the Locomotion Specialist ($\pi_l$).}
\label{tab:reward_pi_l}
\renewcommand{\arraystretch}{1.3}
\resizebox{\linewidth}{!}{
\begin{tabular}{llc}
\toprule
\textbf{Reward Terms} & \textbf{Equation} & \textbf{Weight} \\
\midrule
Track lin. vel. & $\exp\{ -\frac{\|\bm{v}_{\text{lin}}^{\text{cmd}} - \bm{v}_{\text{lin}}\|^2_2}{0.5} \}$ & $5.0$ \\
Track ang. vel. & $\exp\{ -\frac{(\bm{\omega}_{\text{yaw}}^{\text{cmd}} - \bm{\omega}_{\text{yaw}})^2}{0.5} \}$ & $5.0$ \\
Joint acc. & $\|\ddot{\theta}\|_2^2$ & $-5 \times 10^{-7}$ \\
Joint vel. & $\|\dot{\theta}\|_2^2$ & $-1 \times 10^{-3}$ \\
Action rate & $\|\bm{a}_t - \bm{a}_{t-1}\|_2^2$ & $-0.03$ \\
Action smoothness & $\|\bm{a}_t - 2\bm{a}_{t-1}+\bm{a}_{t-2}\|_2^2$ & $-0.05$ \\
Angular vel. ($x, y$) & $\|\bm{\omega}_{xy}\|_2^2$ & $-0.05$ \\
Orientation & $\|\bm{g}_{xy}^\text{torso}\|_2^2$ & $-1.5$ \\
Joint power & $|\tau||\dot{ \theta}|^{\top}$ & $-2.5 \times 10^{-5}$ \\
Feet stumble & $\mathbb{I}(\exists i, |\bm{F}_i^{xy}| \ge 3|F_i^z|)$ & $-1.0$ \\
Torques & $\sum_{\text{all joints}}\tau_i^2$ & $-1 \times 10^{-5}$ \\
Joint deviation & $\sum_{k \in \{ \text{arm, waist, hip} \}} w_k \sum_{j \in \mathcal{J}_k} |\theta_j - \theta_{j}^{\text{def}}|$ & $-0.5$ \\
Joint pos. limits & $\sum_{\text{all joints}}\bm{out}_i$ & $-2.0$ \\
Joint vel. limits & $\text{ReLU}(\dot{\theta} - \dot{\theta}^{\text{max}})$ & $-1.0$ \\
Torque limits & $\text{ReLU}(\tau - \tau^{\text{max}})$ & $-1.0$ \\
Feet lateral distance & $|(y^b_{\text{left feet}} - y^b_{\text{right feet}}) - 0.2|$ & $0.5$ \\
Feet slippage & $\sum_{\text{feet}}\|\bm{v}_i^{\text{foot}}\| \cdot \mathbb{I}_{\text{contact}}$ & $-0.25$ \\
Collision & $n_{\text{collision}}$ & $-15.0$ \\
Feet air time & $\sum_{\text{foot}}(t^{\text{air}}_{i}-0.5) \cdot \mathbb{I}(\text{first contact}_i)$ & $1.0$ \\
Stuck & $(\|\bm{v}\|_2 \le 0.1) \cdot (\|\bm{c}^v\|_2 \ge 0.2)$ & $-1.0$ \\
Feet clearance & $\sum_{\text{foot}} ((z^{i}-h^{\text{target}})^2 \cdot \|\bm{v}^{i}_{xy}\|)$ & $2.0$ \\
Alive & $1$ & $2$ \\
AMP reward & $\displaystyle \text{max}\left[0, 1 - \frac{1}{4}(D_{\phi}(\tau) - 1)^2\right]$ & $3.0$ \\
\bottomrule
\end{tabular}
}
\end{table}

\begin{table}[h]
\centering
\caption{Reward Function Definitions for the Recovery Specialist ($\pi_r$). The $f_{\text{tol}}$ adopts a Gaussian-style formulation, as detailed in \cite{hr:he2025learning}.}
\label{tab:reward_pi_r}
\resizebox{\linewidth}{!}{
\renewcommand{\arraystretch}{1.4}
\begin{tabular}{llc}
\toprule
\textbf{Reward Terms} & \textbf{Equation} & \textbf{Weight} \\
\midrule
\rowcolor{gray!10} \multicolumn{2}{l}{\textit{Task Rewards}} & $w^\text{task}=1.0$ \\
Orientation & $f_{\text{tol}}(-\theta^z_{\text{base}}, [0.99, \infty], 1, 0.05)$ & $1.0$ \\
Head height & $f_{\text{tol}}(h_{\text{head}}, [1, \infty], 1, 0.1)$ & $1.0$ \\ 
\midrule
\rowcolor{gray!10} \multicolumn{2}{l}{\textit{Style Rewards}} & $w^\text{style}=1.0$ \\
Hip joint deviation & $\sum_{\text{hips}}\mathbb{I}(\max|\theta_i|>0.9 \lor \min|\theta_i|<0.8)$ & $-10.0$ \\
Knee deviation & $\sum_{\text{knees}}\mathbb{I}(\max|\theta_i|>2.85 \lor \min|\theta_i|<-0.06)$ & $-0.25$ \\
Shoulder roll dev. & $\mathbb{I}(\theta_\text{left}<-0.02 \lor \theta_\text{right}>0.02)$ & $-2.5$ \\
Thigh orientation & $f_\text{tol}(\frac{1}{2}\sum_\text{thighs}(\theta^{\text{z}}_\text{thigh}), [0.8, \infty], 1, 0.1)$ & $10.0$ \\
Feet distance & $\mathbb{I}(\|\mathbf{p}^{xy}_{\text{left\_f}} - \mathbf{p}^{xy}_{\text{right\_f}}\|^2 > 0.9)$ & $-10.0$ \\
Angular vel. ($x,y$) & $\exp(-2\| \bm{\omega}_{xy}\|^2_2) \cdot \mathbb{I}(h_\text{base} > h_\text{stage1})$ & $25.0$ \\
Foot displacement & $\exp(\text{clip}(-2\| \mathbf{q}^{xy}_\text{base}-\mathbf{q}^{xy}_\text{foot} \|^2, 0.3, \infty))$ & $2.5$ \\ 
AMP reward & $\max\left[0, 1 - \frac{1}{4}(D_{\phi}(\tau) - 1)^2\right]$ & $80.0$ \\
\midrule
\rowcolor{gray!10} \multicolumn{2}{l}{\textit{Regularization Rewards}} & $w^\text{regu}=1.0$ \\
Joint acc. & $\|\ddot{\theta}\|_2^2$ & $-2.5\text{e-}7$ \\
Joint vel. & $\|\dot{\theta}\|_2^2$ & $-1\text{e-}3$ \\
Action rate & $\|\mathbf{a}_t - \mathbf{a}_{t-1}\|_2^2$ & $-0.01$ \\
Action smoothness & $\|\mathbf{a}_t - 2\mathbf{a}_{t-1}+\mathbf{a}_{t-2}\|_2^2$ & $-0.05$ \\
Torques & $\sum \tau_i^2$ & $-1 \times 10^{-5}$ \\
Joint power & $|\tau| |\dot{ \theta}|^{\top}$ & $-2.5 \times 10^{-5}$ \\
Joint pos. limits & $\sum \text{ReLU}(|\theta_i| - \theta_i^{\text{max}})$ & $-2.0$ \\
Joint vel. limits & $\text{ReLU}(\dot{\theta} - \dot{\theta}^{\text{max}})$ & $-1.0$ \\ 
\midrule
\rowcolor{gray!10} \multicolumn{2}{l}{\textit{Post-Task Rewards (Conditioned on $h_\text{base} > h_\text{stage3}$)}}  & $w^\text{task}=1.0$ \\
Tracking errors & $\exp(-2\| \bm{\omega}_{xy}\|_2^2), \exp(-5\| \mathbf{v}_{xy}\|_2^2), \exp(-5\| \mathbf{g}_{xy}\|_2^2)$ & $10.0$ \\
Base height & $\exp(-20| h_\text{base}-0.75|)$ & $10.0$ \\
Target joint dev. & $\exp(-0.1\sum (\theta_i-\theta_i^{\text{def}})^2)$ & $10.0$ \\
Target feet dist. & $f_\text{tol}(y_{\text{L}} - y_{\text{R}}, [0.3, 0.4], 0.1, 0.05)$ & $-5.0$ \\
\bottomrule
\end{tabular}
}
\end{table}

\clearpage

\end{document}